\documentclass[10pt,twocolumn,letterpaper]{article}

\usepackage{wacv}
\usepackage{times}
\usepackage{setspace}
\usepackage{epsfig}
\usepackage{graphicx}
\usepackage{amsmath}
\usepackage{amssymb}
\usepackage{gensymb}
\usepackage{subfigure}
\usepackage{caption}
\usepackage[accsupp]{axessibility}

\def\wacvPaperID{1378} 

\wacvfinalcopy 

\ifwacvfinal
\fi

\ifwacvfinal
\usepackage[breaklinks=true,bookmarks=false]{hyperref}
\else
\usepackage[pagebackref=true,breaklinks=true,colorlinks,bookmarks=false]{hyperref}
\fi

\begin{document}

\title{Attribute-Based Deep Periocular Recognition: \\ Leveraging Soft Biometrics to Improve Periocular Recognition
}

\author{Veeru Talreja, \ Nasser M. Nasrabadi, and \ Matthew C. Valenti\\
West Virginia University\\
Morgantown, WV, USA\\
{\tt\small vtalreja@mix.wvu.edu,nasser.nasrabadi@mail.wvu.edu,valenti@ieee.org }}

\maketitle
\ifwacvfinal
\thispagestyle{empty}
\fi

\begin{abstract}
   In recent years, periocular recognition has been developed as a valuable biometric identification approach, especially in wild environments (for example, masked faces due to COVID-19 pandemic) where facial recognition may not be applicable. This paper presents a new deep periocular recognition framework called attribute-based deep periocular recognition (ADPR), which predicts soft biometrics and incorporates the prediction into a periocular recognition algorithm to determine identity from periocular images with high accuracy. We propose an end-to-end framework, which uses several shared convolutional neural network (CNN) layers (a common network) whose output feeds two separate dedicated branches (modality dedicated layers); the first branch classifies periocular images while the second branch predicts soft biometrics. Next, the features from these two branches are fused together for a final periocular recognition. The proposed method is different from existing methods as it not only uses a shared CNN feature space to train these two tasks jointly, but it also fuses predicted soft biometric features with the periocular features in the training step to improve the overall periocular recognition performance. Our proposed model is extensively evaluated using four different publicly available datasets. Experimental results indicate that our soft biometric based periocular recognition approach outperforms other state-of-the-art methods for periocular recognition in wild environments.



\end{abstract}

\section{Introduction}

Traditionally, facial recognition systems (in cooperative settings) are presented with mostly non-occluded faces, which include all primary facial regions such as the eyes, nose, and mouth \cite{talreja_TIFS,talreja_BTAS_face,talreja2019zero,talreja_icce_2018,taherkhani_talreja_ijcb,taherkhani_talreja_TBIOM,talreja_icc_2019}.  Inspired by the COVID-19 pandemic response, the widespread requirement that people should wear protective face masks in public places has driven a need to understand how cooperative facial recognition technology deals with occluded faces, such as when only the periocular region is visible. In this paper, we address this challenge by implementing a periocular recognition algorithm for unconstrained wild environments (i.e., masked faces).

\begin{figure}[t]
\centering
\captionsetup{justification=centering}
\includegraphics[scale=0.60]{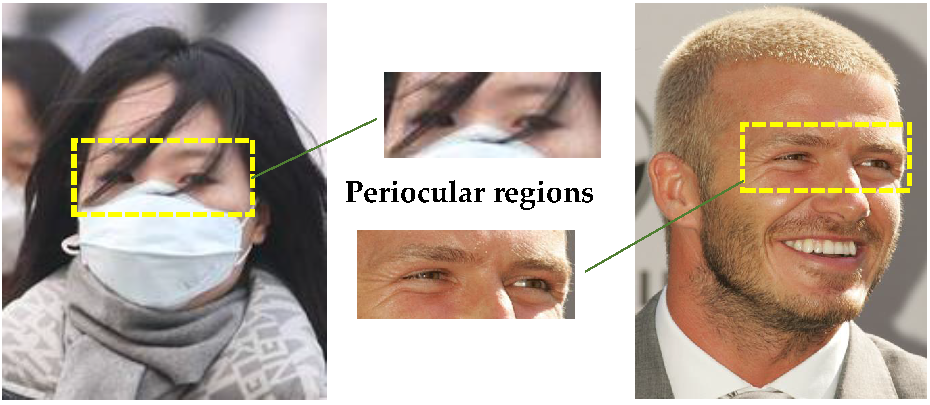}
\caption{Periocular regions \cite{Tiong_AS_2019}.}\label{fig:pr_region}
\end{figure}

In the recent past, periocular recognition in wild environments has garnered significant interest and has become a key area of research in biometric recognition \cite{Park_BTAS_2009,Bharadwaj_BTAS_2010,Padole_ICB_2012,Santos_CIBIM_2013,Tan_TIP_2013,Nie_ICPR_2014,Sharma_ICIP_2014,Smereka_TIFS_2015,Zhao_TIFS_2017,Proenca_TIFS_2018}. Despite the fact that there is no strict definition or standard from professional bodies like ISO/IEC or NIST, the periocular region usually refers to the region around the eye \cite{Zhao_TIFS_2017} as shown in Fig. \ref{fig:pr_region} \cite{Tiong_AS_2019}. The periocular region is considered to be a highly discriminative biometric modality and a powerful alternative/compliment to face and/or iris recognition when accurate face/iris recognition cannot be guaranteed due to the unconstrained environment \cite{Santos_CIBIM_2013} or when the whole face or iris image is not clearly available \cite{Tan_TIP_2013,Woodard_ICPR_2010,Zhao_TIFS_2017}, as illustrated in Fig. \ref{fig:Msked_Faces}. Moreover, it has been shown in the literature that the periocular region is more robust to expression variation \cite{Smereka_TIFS_2015} and aging \cite{Juefei_IJCB_2011}, when compared to other parts of the face. Nonetheless, periocular recognition in wild environments remains a challenging task, largely because the periocular region includes less information than the entire face and is highly susceptible to interference from occlusions (hair and glasses).

When a human recognizes a face, he or she not only analyzes the overall visual pattern but also analyzes semantic information, such as gender, ethnicity, age, etc., to judge whether the face belongs to a certain known person. Therefore, it is reasonable to hypothesize that semantic information is helpful for the task of automated visual identification. In this paper, we propose to use soft biometrics as semantic information for periocular recognition. Soft biometric information extracted from the periocular region of the face (e.g., gender, ethnicity, skin color, and so on) is ancillary information, which to some extent is easily distinguished at a distance but is not fully discriminative by itself during facial or periocular recognition tasks. However, soft biometrics can be explicitly incorporated into periocular recognition algorithms to improve the overall recognition performance when confronting highly variable conditions. We hypothesize that soft biometrics can provide valuable information for periocular recognition, where face images are usually captured in poor quality conditions due to variability in distance, illumination, and pose.  We propose to complement ‘hard’ periocular facial signatures with soft biometrics to improve the overall periocular recognition performance.

We propose a new deep periocular recognition framework called Attribute-based Deep Periocular Recognition (ADPR), which simultaneously predicts soft biometrics and incorporates this ancillary information with a periocular recognition algorithm to determine identity from periocular images with higher accuracy. ADPR is an end-to-end framework, which uses several shared convolutional neural network (CNN) layers (a common network) whose output feeds into two separate branches (modality dedicated layers); the first branch classifies periocular images while the second branch predicts soft biometric attributes. Next, the features from these two branches are fused together for a final periocular recognition. Therefore, in contrast with other existing methods, which only use a shared CNN feature space to train these two tasks jointly, our proposed method fuses the predicted soft biometric features with periocular features in the training step to improve the overall periocular recognition performance.  Thus, our proposed deep model predicts soft biometrics and  simultaneously leverages the predicted soft biometric features as an auxiliary modality to improve periocular recognition performance. In summary, our major contributions are:

\begin{figure}[t]
\centering
\captionsetup{justification=centering}
\includegraphics[scale=0.30]{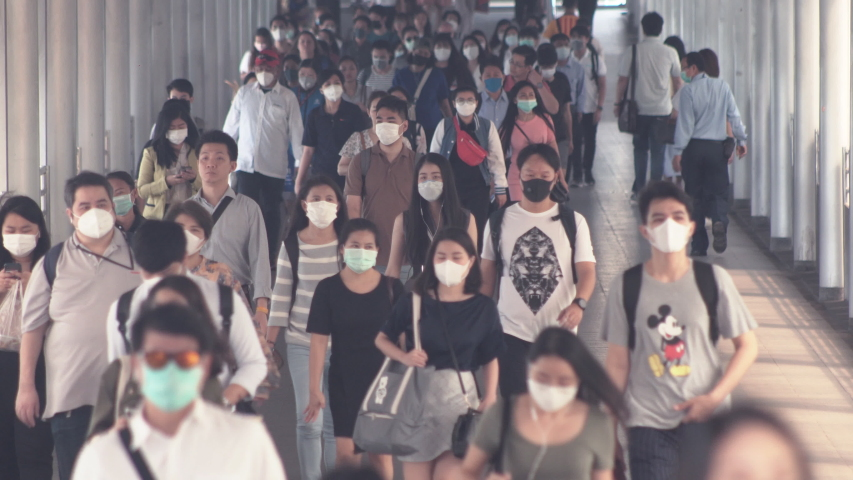}
\caption{Masked Faces.}\label{fig:Msked_Faces}
\end{figure}

\begin{enumerate}
    \item We design and implement an attribute-based deep periocular recognition framework, which is an end-to-end deep periocular recognition algorithm enhanced with the predicted soft biometrics (gender, ethnicity, etc.). 
    
    \item We train the proposed model to effectively predict soft biometrics while simultaneously being trained with the loss from the periocular recognition task. The model shares learned parameters to train both tasks and also fuses soft biometric information with the periocular features to improve the overall periocular recognition performance.
    \item We present an ADPR-based Siamese architecture using contrastive loss, which makes it possible for the ADPR to be used for cross-dataset periocular recognition, which implies that the ADPR does not require training samples from target datasets. 

\end{enumerate}

\section{Related Work}

Periocular recognition in unconstrained conditions has been the subject of ongoing research. Surveys of various periocular recognition algorithms can be found in \cite{Alonso_PRL_2016,Nigam_IF_2015,Rattani_IVC_2017}. A seminal paper investigating the feasibility of using the periocular region for human recognition under various conditions is given by Park \etal \cite{Park_BTAS_2009}. Further research in this area is found in \cite{Bharadwaj_BTAS_2010}, which illustrates the usefulness of periocular recognition when iris recognition fails. There has also been  research on cross-spectrum periocular matching \cite{Sharma_ICIP_2014}, which are based on neural network techniques. 

Further research into periocular recognition has focused on using hand-crafted features \cite{Nie_ICPR_2014,Smereka_TIFS_2015,Smereka_ISBA_2016}. For example, \cite{Nie_ICPR_2014} extracts periocular features using DSIFT and exploits K-means clustering for dictionary learning and representation. However, DSIFT feature extraction and K-mean clustering are computationally expensive and time-consuming. In \cite{Smereka_TIFS_2015}, a Periocular Probabilistic Deformation Model (PPDM) was proposed, which utilizes a probabilistic inference model to evaluate the matching scores from correlation filters on periocular image pairs. The same research group later improved PPDM by selecting discriminative patch regions for more accountable matching \cite{Smereka_ISBA_2016}. However, both of the patch-based matching schemes \cite{Smereka_TIFS_2015,Smereka_ISBA_2016} are less resistant to scale variance among samples that often exists in challenging forensic and security scenarios \cite{Zhao_TIFS_2017}.

Recent developments in periocular recognition techniques are more focused on deep learning-based methods \cite{Zhao_TIFS_2017,Proenca_TIFS_2018}. Proenca and Neves \cite{Proenca_TIFS_2018} proposed Deep-PRWIS, where a deep CNN model is trained in such a way that the recognition is exclusively based on information surrounding the eye, and the iris and sclera regions features are degraded during learning. Zhao and Kumar \cite{Zhao_TIFS_2017} proposed a deep-learning based model called Semantics-Assisted CNN (SCNN), which incorporates explicit semantic information (gender and side) of the training samples to extract more comprehensive periocular features, helping to improve the CNN's performance. 
While the proposed ADPR framework is motivated by SCNN, and similarly leverages soft biometrics for periocular recognition, it additionally fuses soft biometric features with periocular features in a joint learning framework to predict soft biometrics and also improve periocular recognition accuracy.



\section{Proposed Method}

\begin{figure}[t]
\centering
\captionsetup{justification=centering}
\includegraphics[scale=0.35]{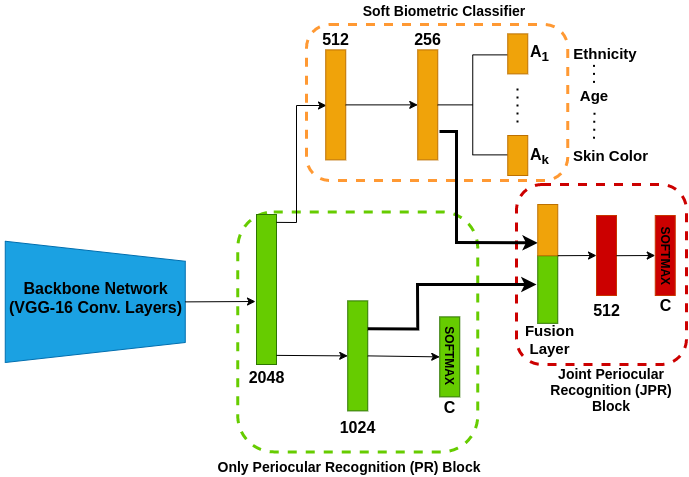}
\caption{Attribute-based Deep Periocular Recognition (ADPR). The rectangular solid colored boxes with numbers depict fully connected layers. }\label{fig:arch}
\vspace{-0.25cm}
\end{figure}

The proposed end-to-end deep learning framework is shown in Fig. \ref{fig:arch}. It consists of four important blocks: The backbone network, the only periocular recognition (PR) block (green dashed section), the soft biometric classifier (orange dashed section), and the joint periocular recognition (JPR) block (red dashed section). Specifically, the architecture uses several CNN layers for the backbone network whose output feeds into two separate branches (modality specific layers); the first branch is for only periocular recognition while the second branch is for soft biometric classification. The features from the soft biometric classifier and the PR block are fused in the JPR block for a final joint periocular recognition.

The backbone network is formed by only the convolutional layers of VGG-16 \cite{Simonyan_ICLR_2015} pretrained on ImageNet \cite{deng_CVPR_2009}. ResNet\cite{He_CVPR_2016} can also be used in place of VGG-16 for the backbone network. For the PR block and soft biometric classifier, a new set of fully connected layers (green and orange dashed sections) are integrated with the backbone network. During optimization, the periocular layers (green) are optimized for only periocular recognition followed by optimization of the soft biometric layers (orange) for only soft biometric prediction.

For the JPR block, the features from the periocular layers are fused with the features from the soft biometric layers (shown in bold black arrows) in the fusion layer followed by additional fully connected layers and a Softmax layer for periocular recognition again. However, this time the periocular recognition uses features from both the soft biometric layers and the periocular layers. This is why it is called the \emph{joint} periocular recognition block. Finally, the whole network (except the Softmax layer in PR block) is trained end-to-end for joint periocular recognition and soft biometric classification, and the loss from both the joint periocular recognition and soft biometric classification is back propagated through the network. The novelty is that, because of this feature fusion and loss propagation, the soft biometrics features enhance the discriminative power of the periocular recognition network and therefore improve the overall periocular recognition performance. The implementation details for each block of the proposed architecture are defined in the following subsections.

\subsection{Backbone Network}
The backbone network uses the 14 convolutional layers from the VGG-16 architecture pre-trained on the ImageNet dataset as a starting point. The fully connected layers of the VGG-16 are discarded and not part of the backbone network. There are multiple reasons for using the VGG-16 pre-trained on the ImageNet dataset for the backbone network. In the proposed architecture, the backbone network is only used as feature-extractor for periocular recognition and soft biometric classification, and it can be seen from the previous literature \cite{Nguyen_2017_IrisRW,zhao_2017_towards,minaee_2016_experimental,Schroff_2015_FaceNetAU,Sun_2015_DeepID3,Parkhi_2015_DeepFR} that the features provided by VGG-16 pre-trained on ImageNet and fine-tuned on different biometric image data are very discriminative and therefore can be used as a starting point. Moreover, starting with a well-known architecture makes the work highly reproducible.
\subsection{Only Periocular Recognition Block}
For the PR block, a new set of fully connected layers (green blocks in the green dashed section) for only periocular recognition are integrated with the backbone network. Specifically, we add two fully connected layers with the first layer having 2,048 nodes and the second layer having 1,024 nodes. The final layer in the PR block is the Softmax layer, where the size of the layer depends on the number of classes \emph{C} in the given dataset.

\subsection{Soft Biometric Classifier}
In addition to improving periocular recognition, another important objective of our proposed method is effective prediction of soft biometrics using only the periocular region. However, separating these two objectives by training multiple CNNs individually is not optimal since different objectives may share common features and have hidden relationship, which can be leveraged to jointly optimize the objectives. This notion of joint optimization has been used in \cite{Zhong2016FaceAP}, where they train a CNN for face recognition, and utilize the features for attribute prediction. 

Therefore, for this task, we use the feature set output by the first fully connected layer (size 2,048) of the PR block to also predict the soft biometrics for a given periocular image. For the soft biometric classifier we add two fully connected layers, a first layer of size 512 and a second layer of size 256, as shown in Fig. \ref{fig:arch}. The final layer in the soft biometric classifier is the set of $k$ binary classifiers, where $k$ denotes the number of soft biometrics predicted.


\subsection{Joint Periocular Recognition}

After optimizing the soft biometric classifier and PR block individually, the features from the soft biometric layers are fused with the features from the periocular layers at the fusion layer of the JPR block. The fusion layer is followed by additional fully connected layers and a Softmax layer for the final periocular recognition. Thus, the final periocular recognition utilizes the fused features from the periocular layers and the soft biometric layers. The whole system is then trained end-to-end and the loss from both the joint periocular recognition block and soft biometric classification is back propagated through the network. 

In our implementation, we use the output features from the second fully connected layer (size 256) of the soft biometric classifier and vertically concatenate it with the features of the second fully connected layer (size 1,024) of the PR block. This vertical concatenation is the fusion layer. The output of the fusion layer is of size 1,280. In addition to this fusion layer, we add another fully connected layer of size 512 followed by a Softmax layer, whose output is equal to the number of classes \emph{C} in the given dataset.
\vspace{-0.20cm}
\section{Training of the Proposed Architecture}

The proposed architecture is trained in three steps:

1. In the first step, only the PR block is optimized for the periocular classification of the input images. Consider that the input image to ADPR is a periocular image denoted by $x^i$, where the class label for the image is given by $y^{i} \in C$ for $i=1,\cdots,N$ where $N$ is the number of training images in a mini-batch. In this step, we only optimize the PR block's layers for classification while keeping the backbone network frozen.  The classification formulation has been incorporated into the PR block by adding the Softmax layer as shown in Fig. \ref{fig:arch}. Let $E_{1}$ denote the objective function required for classification:
    \vspace{-0.25cm}
\begin{equation} E_{1}(\textbf{w}_{PR})=\frac{1}{N}\sum_{i=1}^{N}L_i(f(x^{i},\textbf{w}_{PR}),y^{i}) + \lambda ||\textbf{w}_{PR}||^{2} \label{eq:1},\end{equation} where the first term $L_i(.)$ is the classification loss for training instance $i$ and $N$ is the number of training images in a mini-batch. $f(x^{i},\textbf{w}_{PR})$ is the predicted softmax output of the PR network and is a function of the input training image $x^i$ and the weights of the PR network $\textbf{w}_{PR}$. The last term is the regularization function where $\lambda$ governs the relative importance of the regularization. 

The choice of the loss function $L_i(.)$ depends on the application. We use a classification loss function that uses softmax outputs to minimize the cross-entropy error function. Let the predicted softmax output $f(x^{i},\textbf{w}_{PR})$ be denoted by $\hat{y}^{i}$. The classification loss for the $i^\mathsf{th}$ training instance is given as:
\vspace{-0.25cm}
\begin{equation}L_{i}(\hat{y}^{i},y^{i})=-\sum_{m=1}^{C}{y}^{i}_{m}\ln \hat{y}^{i}_{m} \label{eq:2},\end{equation} where $y^{i}_{m}$ and $\hat{y}^{i}_{m}$ are the ground truth and the prediction result for the $m^\mathsf{th}$ softmax output of the $i^\mathsf{th}$ training instance, respectively and $C$ is the number of softmax outputs.

2. In the second step, only the soft biometric classifier block is optimized for the prediction of the soft biometrics.  For the soft biometric prediction task, a periocular image is given as input to predict a set of soft biometrics. The periocular image  is again denoted by $x^i$, the number of different soft biometrics being predicted is denoted by $k$, and $A^{i}_{t}$ denotes the ground truth soft biometric label for training sample $i$ and attribute $t$ for $t=1, \cdots, k$. In this case, using the feature set from the second fully connected layer (size 256), the soft biometric prediction loss function is denoted by $E_{2}$, and is given as:
\vspace{-0.25cm}
 \begin{equation}
     \begin{split}
         E_{2}(\textbf{w}_{t})=\frac{1}{N}\sum_{i=1}^{N}\sum_{t=1}^{k}L^{i}_{t}(f^t(z_1(x^i)\times \textbf{w}_{t}),A^{i}_{t}),
     \end{split}\label{eq:3}
 \end{equation}
where $f^t(.)$ is a binary classifier for the soft biometric $t$ operating on the feature set from the second fully connected layer (size 256) of the soft biometric classifier. The classifier is learned by using a binary cross-entropy loss function $L^i_t(.)$. $\textbf{w}_{t}$ represents the weight parameters for the classifier, and these parameters are learned separately for each soft biometric attribute. 

3. In the final step, the JPR block and the soft biometric classifier are optimized together. The loss function used to train the network in this step is a combination of $E_1$ and $E_2$, which implies it is a combination of classification loss and soft biometric prediction loss functions. Let $E_3$ denote this combination given as:
\vspace{-0.15cm}
\begin{equation}
     \begin{split}
        E_{3}(\textbf{w}_{JPR},\textbf{w}_{t})= E_1(\textbf{w}_{JPR}) + E_{2}(\textbf{w}_{t}),
     \end{split}\label{eq:4}
 \end{equation}
 where $\textbf{w}_{JPR}$ signifies that the classification loss formulation $E_1$ is a function of the JPR block layers which includes the fused feature vector of the soft biometric classifier features and the periocular recognition features. Finally, the whole network (except the Softmax layer in the PR block) is trained end-to-end using the $E_3$ loss function.
 
For each training step, we have used the stochastic gradient descent (SGD) algorithm,
with a batch size of 64 samples. The learning rate was $10^{-3}$, with a momentum of 0.9 and a weight decay of $5\times 10^{-4}$ .  

 \section{Open vs. Closed-World Setting}
 
 When CNNs are used for recognition, it is important to understand and decide if the implemented system is expected to work in an open-world or closed-world mode; i.e., if the system has access to all the training time samples from all the classes that will be seen at the testing time or not. In a scenario where the CNNs are trained in a classification protocol (i.e., the identity or the category of the input data is known), the closed-world mode can be enabled and the output of the nodes in the final Softmax layer can be used as probabilities for each class label. However, in the open-world mode, one-to-one matching for probably unseen subjects is the key problem and needs to be evaluated. Therefore, in the open-world setting, the model needs to be trained to be able to generalize to unseen subjects that are not included in the training set. In this paper, we have evaluated our proposed architecture for both open-world and closed-world settings. 


\subsection{Siamese Network for Open-World Setting} 
\begin{figure}[t]
\centering
\captionsetup{justification=centering}
\includegraphics[scale=0.55]{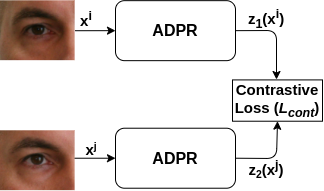}
\caption{ADPR-based Siamese}\label{fig:siamese_arch}
\vspace{-0.25cm}
\end{figure}

For the open-world setting, we use a Siamese network \cite{Chopra_CVPR_2005} to train the ADPR, generalize to unseen subjects, and predict the similarity between a pair of feature vectors. The Siamese network is primarily designed for verification scenarios. The ADPR-based Siamese network architecture is shown in Fig. \ref{fig:siamese_arch}. The Siamese network requires genuine and impostor pairs of images. In the Siamese architecture, we use identical ADPR networks with shared weights for both the input images, and these two networks are coupled together using a contrastive loss function $(L_{cont})$. We use the output of the fully connected layer (size 512) of the JPR block in ADPR as the feature representation of the input data and for coupling the two networks using $L_{cont}$. While the Softmax layer in the JPR block represents the class probabilities during the training process, the second-to-last layer should contain the most relevant and aggregated information that can contribute to distinguishing the classes or identities \cite{Zhao_TIFS_2017}. Therefore, it is reasonable to use the output of the fully connected layer (size 512) of the JPR block as the feature representation and generalize the model to unseen subjects. 

The loss function  $L_{cont}$ is minimized so as to drive the genuine pairs (i.e., both the periocular images belonging to the same subject) towards each other in the feature domain, and at the same time, push the impostor pairs (i.e., the periocular images belonging to different subjects) away from each other. Let $x^i$ denote the one input periocular image, and $x^j$ denote the other input periocular image as shown in Fig. \ref{fig:siamese_arch}. Define $c(i,j)$ to be a binary label, which is equal to 0 if $x^i$ and $x^j$ belong to a genuine pair, and equal to 1 if $x^i$ and $x^j$ belong to an impostor pair. Let $z_1(.)$ and $z_2(.)$ denote the ADPR network to transform  $x^i$ and $x^j$, respectively into their feature representation of size 512. If $c(i,j)=0$ (i.e., genuine pair), the contrastive loss function is:
\begin{equation}
\begin{split}
L_{cont}(z_1(x^i),z_2(x^j),c(i&,j)) = \frac{1}{2}\left\lVert z_1(x^i)-z_2(x^j)\right\rVert^2_2.  
  \end{split}
  \end{equation}\label{eq:5}Similarly if $c(i,j)=1$, then contrastive loss function is :
  \begin{equation}
  \begin{split}
L_{cont}(z_1(x^i),z_2(&x^j),c(i,j))  = \\ & \frac{1}{2}\mbox{max}\biggl(0,m-\left\lVert z_1(x^i)-z_2(x^j)\right\rVert^2_2\biggr), 
\end{split}
\end{equation}\label{eq:6}where $m$ is the contrastive margin used to ``tighten" the constraint. Therefore, the total loss function for training the Siamese architecture is denoted by $L_{cpl}$ and is given as: 
\begin{equation}
    \begin{split}
        L_{cpl}=\frac{1}{N^2}\sum_{i=1}^{N}\sum_{j=1}^{N}L_{cont}(z_1(x^i),z_2(x^j),c(i,j)), 
    \end{split}\label{eq:7}
\end{equation}
where $N$ is the number of training samples. The main motivation for using the coupling loss is that it has the capacity to find the discriminative embedding subspace because it uses the class labels implicitly, which may not be the case with some other metrics such as the Euclidean distance.

Our approach for periocular recognition in the open-world setting using the trained Siamese ADPR network does not require that the training and testing samples belong to the same dataset, which is a key advantage over other state-of-the-art (SOTA) approaches \cite{Tan_TIP_2013,Proenca_TIFS_2018,Smereka_TIFS_2015}. In our experiments for the open-world setting, the ADPR is trained with one database and tested on a separate dataset. The testing and training sets have mutually exclusive subjects and highly different image qualities, conditions, and/or sensors.



 \section{Performance Evaluation}
 In this section, we discuss the datasets and the performance evaluation of our proposed architecture when compared to other state-of-the-art methods. For fair comparison, we have maintained consistency with \cite{Proenca_TIFS_2018}  and \cite{Zhao_TIFS_2017} with respect to the dataset and the testing protocol. 
 \subsection{Datasets}
 We have used the following publicly available databases for the experiments:

1) \textbf{UBIPr} \cite{Padole_ICB_2012}: We have used the UBIPr database only for training our proposed ADPR architecture in the open-world setting. Originally, this database contains 5,126 images for each of the left and right perioculars from 344 subjects. However, following the protocol in \cite{Zhao_TIFS_2017}, we have removed those subjects that are also in UBIRISV2 \cite{Proenca_PAMI_2010}. Finally, for each of the left and right perioculars, we end up with 3,359 images from 224 subjects.

2) \textbf{UBIRISV2} \cite{Proenca_PAMI_2010}: The UBIRISV2 database contains periocular images, and is mainly used for assessment of at-a-distance iris recognition algorithms under visible illumination and challenging imaging environments \cite{Zhao_TIFS_2017}. We have used this dataset to evaluate the performance of both open-world and closed-world settings. For the open-world setting, using the protocol defined in \cite{Zhao_TIFS_2017}, we have used a subset of about 1,000 images from this dataset for the performance evaluation. This subset contains both left and right periocular images corresponding to $161$ subjects, that are captured from a distance of 3 to 8 meters. For the closed-world setting, we have made our dataset protocol consistent with \cite{Proenca_TIFS_2018} and used all 11,100 images corresponding to $522$ different eyes for the training and evaluation of the proposed architecture.  The UBIPr and UBIRISV2 datasets are annotated with only gender as a soft biometric.  

3) \textbf{Face Recognition Grand Challenge (FRGC)} \cite{Phillips_CVPR_2005}: The FRGC dataset consists of full facial images. We have used Multi-task Cascaded Convolutional Networks (MTCNN) \cite{Zhang_MTCNN_2016} to crop out left and right periocular images. We have used the FRGC dataset for both open-world and closed-world settings. For the open-world setting, consistent with \cite{Zhao_TIFS_2017}, we have used only right periocular images from the Fall 2002 subset with subject IDs from 202-269 to 202-317. This corresponds to about 540 right eye images from 163 subjects. For the closed-world setting, we have used the Spring 2004 subset which consists of about 25,000 images corresponding to $690$ classes. The FRGC dataset is annotated with only gender and ethnicity as soft biometrics.    

For open-world setting experiments, it is important to clarify the difference in the training techniques for four methods (Our proposed approach and the three comparable approaches from \cite{Zhao_TIFS_2017,Smereka_TIFS_2015,Tan_TIP_2013}).  For our approach and \cite{Zhao_TIFS_2017}, the model is trained on the UBIPr database and tested on the UBIRISV2 and FRGC databases, which is identical to the train/test configuration in \cite{Zhao_TIFS_2017} and therefore provides a fair comparison. The other two methods \cite{Smereka_TIFS_2015,Tan_TIP_2013} require within-database training and testing and this offers better results for these two methods. Therefore, the training and testing are performed on the same dataset for them. Additionally, for fair comparison, the train/test configuration for these two methods is similar to the configuration in \cite{Zhao_TIFS_2017}.

\subsection{Open-World Performance}
\begin{figure*}[t]
\centering     
\captionsetup{justification=centering}
\subfigure[UBIRISV2]{\label{fig:UBIRISV2_open}\includegraphics[scale=0.52]{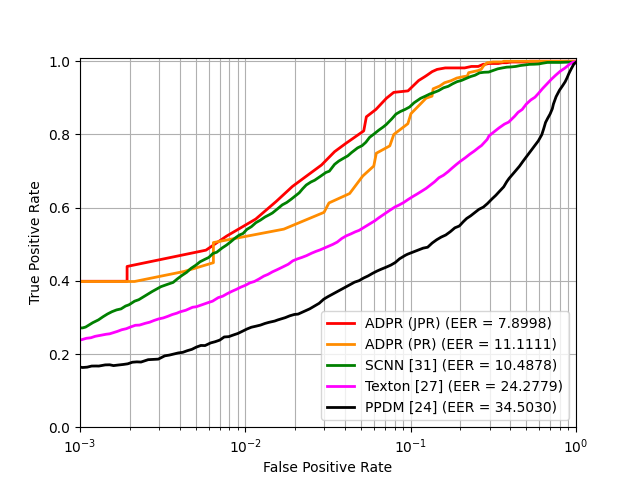}}
\subfigure[FRGC]{\label{fig:FRGC_open}\includegraphics[scale=0.52]{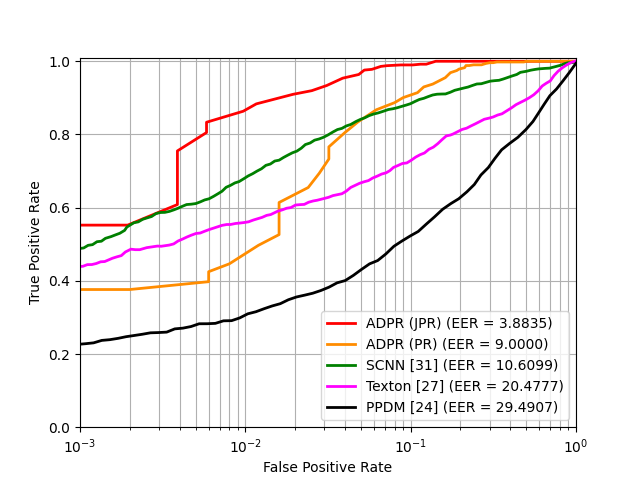}}
\caption{ROC curve comparison for open-world setting against the output of PR block and other state-of-the-art methods for two different datasets.}
\label{fig:ROC_open}
\end{figure*}

We first examine the advantage of the fusion of soft biometric features and periocular recognition features in the JPR block. We have compared the performance of a periocular recognition at the output of the PR block (ADPR (PR)) with the performance of a periocular recognition at the output of the JPR block (ADPR (JPR)). We have used the proposed Siamese architecture with contrastive loss, which is designed for open-world verification. The model is trained on the UBIPr dataset and tested on the UBIRISV2, and FRGC datasets. When training with contrastive loss, the margins are discretely tuned in the range [0.5, 5],  and the results providing best performance are used for comparison. The results from the verification experiments using receiver operating characteristic (ROC) curves are illustrated in Fig. \ref{fig:ROC_open}. It can be observed that the proposed ADPR (JPR) consistently outperforms the output of the PR block (ADPR (PR)). This observation suggests that, due to the feature fusion and back propagation of the combined loss function, the soft biometrics are directly influencing the discriminative power of the network and therefore improving the overall periocular recognition performance.

We also compared the performance of our approach with other SOTA approaches \cite{Zhao_TIFS_2017,Smereka_TIFS_2015,Tan_TIP_2013} on the periocular recognition problem. We have made the test protocols consistent with \cite{Zhao_TIFS_2017}. The verification results (ROC) for these comparisons are also shown in Fig. \ref{fig:ROC_open}. It can be observed from Fig. \ref{fig:ROC_open} that the proposed approach using ADPR consistently outperforms the three SOTA approaches; SCNN \cite{Zhao_TIFS_2017}, PPDM \cite{Smereka_TIFS_2015}, Texton \cite{Tan_TIP_2013}. For instance, in case of the UBIRISV2 dataset, the EER for our proposed ADPR (JPR) is improved by $3\%$ over the second best method SCNN \cite{Zhao_TIFS_2017}. 
\subsection{Closed-World Performance}

\begin{figure*}[t]
\centering     
\captionsetup{justification=centering}
\subfigure[UBIRISV2]{\label{fig:UBIRISV2_closed}\includegraphics[scale=0.52]{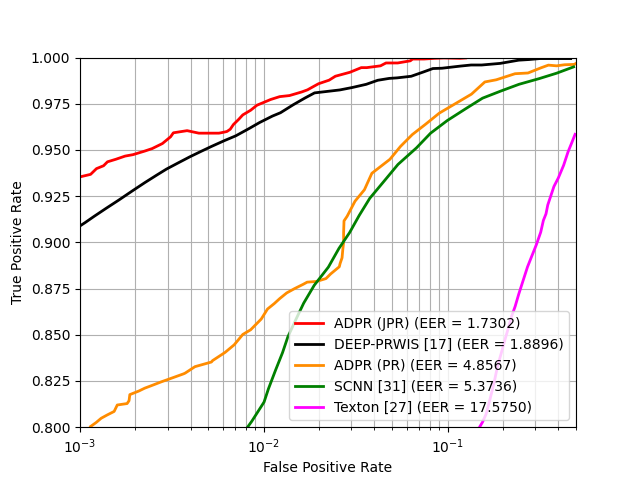}}
\subfigure[FRGC]{\label{fig:FRGC_closed}\includegraphics[scale=0.52]{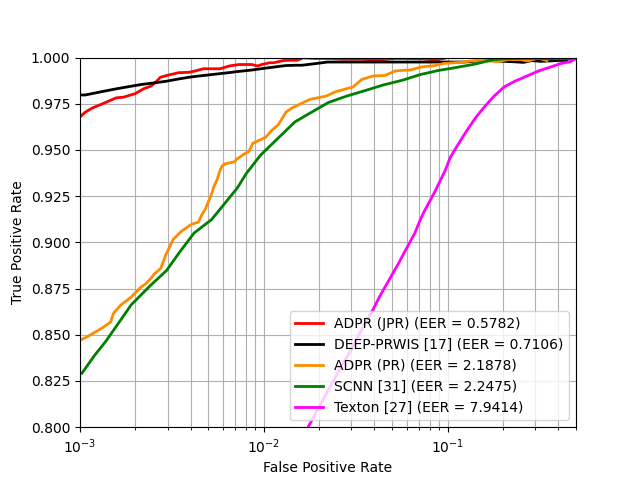}}
\caption{ROC curve comparison for closed-world setting against the ADPR (PR) and other state-of-the-art methods.}
\label{fig:ROC_closed}
\end{figure*}

As discussed earlier, the proposed Siamese ADPR network is primarly designed for open-world verification. However, some recent techniques \cite{Rattani_ICIP_2016,Proenca_TIFS_2018} also focus on the closed-world setting, where there is an overlap of subjects during training/testing phase. Under the closed-world setting, we have maintained the ADPR architecture from Fig. \ref{fig:arch}.  The size of the Softmax layer $C$ is consistent during training and test phases in this closed-world setting. The output for each neuron of the Softmax layer is considered the probability that the input sample belongs to a specific subject, and therefore is used as the verification score. Fig. \ref{fig:ROC_closed} provides ROCs for the verification results under the closed-world setting on UBIRISV2 and FRGC, and includes a comparison to the performance when using only the PR block (ADPR (PR)), and three other SOTA methods. 

From Fig. \ref{fig:ROC_closed}, we observe that our approach consistently outperforms the recent SOTA method DEEP-PRWIS \cite{Proenca_TIFS_2018}. Under the closed-world settings, our results have scored significantly low EER
($1.73\%$). This is due to the fact that, with the feature fusion and back propagation of the combined loss function, the soft biometric features enhance the discriminative power of the network and therefore improve the overall periocular recognition performance.



We have also evaluated the Area Under the Curve (AUC), Rank-1 accuracy, and Equal Error Rate (EER) of our proposed system and compared it against the earlier SOTA methods \cite{Zhao_TIFS_2017,Tan_TIP_2013,Proenca_TIFS_2018}.  
Table \ref{table:perf_closed} summarizes the performance observed in our experiments, for the three algorithms and two data sets considered. The proposed method outperforms the earlier methods, with the true identity being reported at the first position (Rank-1) over $92.68\%$ of the time. In all performance measurements, the differences with respect to the second-best method \cite{Proenca_TIFS_2018} is evident.  

\begin{table}[t]
\centering
\caption{Performance comparison between proposed method and other sate-of-the-art methods for closed-world setting. Values given are percentages.}
\scalebox{0.75}{\begin{tabular}{c c c c }
 \hline
\multicolumn{1}{c}{Method} &\multicolumn{1}{c}{AUC} &\multicolumn{1}{c}{Rank-1}&\multicolumn{1}{c}{EER}\\ [0.5ex] 
 \hline \hline
 &\multicolumn{2}{c}{UBIRISV2}&\\
 ADPR (JPR) & $\boldsymbol{99.9\pm0.02}$&$\boldsymbol{92.68\pm1.04}$&$\boldsymbol{1.73\pm0.04}$\\
  ADPR (PR) & $98.2\pm0.07$&$83.95\pm0.87$&$4.85\pm0.12$\\
 DEEP-PRWIS \cite{Proenca_TIFS_2018} & $99.8\pm0.04$&$87.64\pm1.68$&$1.89\pm0.04$\\
 SCNN \cite{Zhao_TIFS_2017} & $98.6\pm0.05$&$79.3\pm1.6$&$5.4\pm0.04$\\
 Texton \cite{Tan_TIP_2013} & $84.3\pm0.09$&$64.6\pm2.3$&$17.57\pm0.32$\\
 \hline \hline
 &\multicolumn{2}{c}{FRGC}&\\
 ADPR (JPR) & $\boldsymbol{99.9\pm0.01}$&$\boldsymbol{93.65\pm0.88}$&$\boldsymbol{0.58\pm0.02}$\\
  ADPR (PR) & $98.7\pm0.04$&$88.25\pm0.87$&$2.19\pm0.09$\\
 DEEP-PRWIS \cite{Proenca_TIFS_2018} & $99.9\pm0.04$&$92.05\pm0.92$&$0.71\pm0.02$\\
 SCNN \cite{Zhao_TIFS_2017} & $99.1\pm0.04$&$89.2\pm0.79$&$2.25\pm0.04$\\
 Texton \cite{Tan_TIP_2013} & $94.1\pm0.05$&$73\pm2.1$&$7.9\pm0.45$\\
 \hline


\end{tabular}}
\label{table:perf_closed}
\end{table}





\subsection{Soft Biometric Prediction}

In this section, we evaluate the effect of fusing the soft biometric features with the periocular recognition features on periocular recognition performance and on soft-biometric prediction performance. In this experiment, we evaluated the soft biometric prediction performance using classification accuracy before the fusion of the features (ADPR (SB)) and also after the feature fusion with end-to-end model training (ADPR (JPR)). This experiment has been performed on the UBIRISV2 and the FRGC datasets under the closed-world setting. Additionally, for the FRGC dataset we have compared the performance with another earlier method based on LBP features \cite{Lyle_BTAS_2010}. 

Table \ref{table:soft_biom_perf} provides the comparisons of gender accuracy and ethnicity accuracy for the different models discussed above. It can be observed from Table \ref{table:soft_biom_perf} that after feature fusion, there is an improvement of at least $1.5\%$ in soft biometric prediction accuracy. This improvement can be attributed to the feature fusion and end-to-end model training of the ADPR, where the periocular recognition features enhance the soft biometric discriminative space and therefore improve the soft biometric prediction accuracy.


\begin{table}[t]
\centering
\caption{Soft biometric prediction performance comparison.}
\scalebox{0.90}{\begin{tabular}{c c c }
 \hline
\multicolumn{1}{c}{Method} &\multicolumn{1}{c}{Gender Acc.} &\multicolumn{1}{c}{Ethnicity Acc.}\\ [0.5ex] 
 \hline \hline
 &\multicolumn{2}{c}{UBIRISV2}\\
 ADPR (JPR) & $\boldsymbol{97.6}$&-\\
  ADPR (SB) & $95.76$&-\\
 \hline \hline
 &\multicolumn{2}{c}{FRGC}\\
  ADPR (JPR) & $\boldsymbol{97.5}$&$\boldsymbol{98.7}$\\
  ADPR (SB) & $96.1$&$97.1$\\
  LBP \cite{Lyle_BTAS_2010} & $92.5$ & $94.5$\\ 
 \hline
\end{tabular}}
\label{table:soft_biom_perf}
\end{table}


\subsection{Effect of the Number of Soft Biometrics on Periocular Recognition}

\begin{figure}[t]
\centering
\captionsetup{justification=centering}
\includegraphics[scale=0.55]{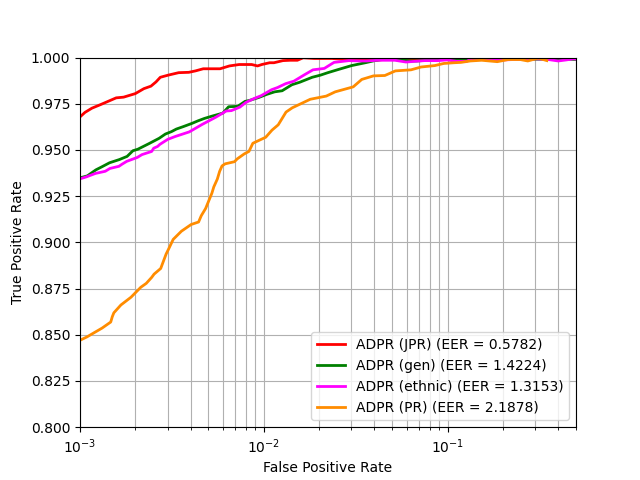}
\caption{Performance comparison for different number of soft biometrics.}\label{fig:roc_perf_diff_soft_biom}
\vspace{-0.25cm}
\end{figure}

We have also evaluated the effect of the number of soft biometrics being used on the periocular recognition performance. We have performed this experiment on the FRGC dataset under a closed-world setting. For this experiment, we have trained the ADPR model using only one soft biometric (gender or ethnicity) and tested it for verification.  We have compared the performance of no soft biometric fusion (ADPR (PR)), only the gender being used as the soft biometric (ADPR (gender)), only the ethnicity being used as the soft biometric (ADPR (ethnicity)), and both gender and ethnicity being used (ADPR (JPR)). Fig. \ref{fig:roc_perf_diff_soft_biom} provides the ROC curve comparison for different models. It can be observed from Fig. \ref{fig:roc_perf_diff_soft_biom} that fusion of soft biometrics with the periocular features helps to improve the overall performance and also that using more soft biometric attributes improves the performance. This is because with more soft biometrics, it becomes easier for the network to the learn the discriminative space, which improves the overall performance.

\section{Conclusion}
We have presented a periocular recognition framework based on a convolutional neural network (CNN) architecture and the fusion of soft biometric features with periocular features. The utility of this framework is that, due to the fusion of soft biometrics and periocular features, along with end-to-end model training, the soft biometric features enhance the discriminative power of the network and therefore improve the overall periocular recognition performance. We observed an improvement in EER of at least $3\%$ in the open-world setting verification performance and an improvement in Rank-1 accuracy of at least $2\%$ in the closed-world setting, when compared to the state-of-the-art methods. We have also evaluated the soft biometric prediction performance and observed an improvement of at least of $1.5\%$ in accuracy due to the fusion of periocular features with the soft biometric features.      

{\small
\bibliographystyle{ieee_fullname}
\bibliography{egpaper}
}

\end{document}